\documentclass[runningheads]{llncs}

\usepackage[T1]{fontenc}

\usepackage{xurl}
\usepackage{graphicx}
\graphicspath{ {images/} }

\begin{document}

\title{Talking like one of us: Effects of using regional language in a Humanoid Social Robot}

\titlerunning{Effects of using regional language in a Humanoid Social Robot}

\author{Thomas Sievers\orcidID{0000-0002-8675-0122} \and
Nele Russwinkel\orcidID{0000-0003-2606-9690}}

\institute{Institute of Information Systems, University of Lübeck, 23562 Lübeck, Germany
\email{sievers@uni-luebeck.de, russwinkel@uni-luebeck.de}}

\maketitle
%\thispagestyle{empty}
%\pagestyle{empty}

%%%%%%%%%%%%%%%%%%%%%%%%%%%%%%%%%%%%%%%%%%%%%%%%%%%%%%%%%%%%%%%%%%%%%%%%%%%%%%%%
\begin{abstract}
Social robots are becoming more and more perceptible in public service settings. For engaging people in a natural environment a smooth social interaction as well as acceptance by the users are important issues for future successful Human-Robot Interaction (HRI). The type of verbal communication has a special significance here. In this paper we investigate the effects of spoken language varieties of a non-standard / regional language compared to standard language. More precisely we compare a human dialog with a humanoid social robot Pepper where the robot on the one hand is answering in High German and on the other hand in Low German, a regional language that is understood and partly still spoken in the northern parts of Germany. The content of what the robot says remains the same in both variants. We are interested in the effects that these two different ways of robot talk have on human interlocutors who are more or less familiar with Low German in terms of perceived warmth, competence and possible discomfort in conversation against a background of cultural identity. To measure these factors we use the Robotic Social Attributes Scale (RoSAS) on 17 participants with an age ranging from 19 to 61. Our results show that significantly higher warmth is perceived in the Low German version of the conversation.
\end{abstract}


%%%%%%%%%%%%%%%%%%%%%%%%%%%%%%%%%%%%%%%%%%%%%%%%%%%%%%%%%%%%%%%%%%%%%%%%%%%%%%%%
\section{INTRODUCTION}

Social robots are coming into more and more contact with people in everyday life. Therefore, it is becoming increasingly important for human-robot interaction (HRI) to find ways for acceptance, good cooperation and collaboration.

In their research van Pinxteren et al. \cite{vanPinxteren} focus on anthropomorphism as a central concept in HRI. According to theory, human-like features of the robot facilitate anthropomorphism \cite{Epley}. And more than 80 per cent of respondents surveyed by TU Darmstadt on the robotization of office and service professions believe that robots can show feelings \cite{StockHomburg}. So people tend to think of robots as social beings. This suggests that for a robot to be accepted by people as something to be involved with, it is important to have a suitable personality and appropriate manners.

Winkle et al. expect carefully designed \textit{social identities} in social robots for maximized effectiveness \cite{Winkle}. They conclude that the deliberate use of anthropomorphic social behaviour in a social robot is essential for good functioning and carries relatively low ethical risks. It is therefore important that a robot addresses people and engages with them in an appropriate manner, not least because many people are still surprised that they can communicate with a robot in a natural way using natural language \cite{Gardecki}. Robots will be most successful in being accepted if they meet expectations through aesthetic characteristics that engage users within a clearly defined context \cite{Diana}. Everything from the robots appearance to the visual design and the sound of voice hold semantic value. Previous work has also provided insights into the effects of language and cultural context on the credibility of robot speech \cite{Andrist}. The way robots express themselves to build credibility and convey information in a meaningful and compelling way is a key function in creating acceptance and usability.

In this paper we investigate the effects of spoken language varieties of a non-standard / regional language compared to standard language. We focus on the effects that these two variations of robot talk have on human interlocutors in terms of perceived warmth, competence and a discomfort in conversation. The regional language provides a cultural identity and serves as a variable for examining the influence in the perception of the robot's personality.

\section{PRELIMINARIES}

The results of a survey conducted by Foster et al. \cite{Foster} concerning expectations for conversational interaction with a robot confirm the significance of accent and dialect. They aim to develop a robot capable of fluid natural-language conversations in socially and ethnically-diverse areas. The circumstances regarding the differences in accent, grammar and vocabulary in their study environment in Glasgow, Scotland, between Scottish English and standard language seem to be similar to those in our environment with Low German and High German.

Low German is a West Germanic language variety spoken mainly in northern Germany and the northeastern part of the Netherlands. It is closely related to Frisian and English, with which it forms the North Sea Germanic group of the West Germanic languages. Low German has not undergone the High German consonant shift. There are about 1.6 million speakers in Germany, mainly in northern Germany. Low German is part of the cultural heritage of northern Germany and complements some of the characteristics and stereotypes people have in mind when they think of a typical northern German person, such as taciturnity, sarcasm and aloofness.

So our research question is to what extent the human perception of a conversation with a social robot speaking in Low German differs from the perception when the robot speaks the standard language regarding warmth, competence and discomfort. Linguistic variation in cultural context has not been well studied in human-robot interaction to the best of our knowledge. With our work we want to take a step forward on this path.

\section{RELATED WORK}

As research has shown, people tend to treat machines like robots and computers as real social beings \cite{Nass}. This behavior includes gender stereotypes and social role models. Nias et al. discuss how speech and dialect shape one's identity and state the importance of a culturally rich engagement experience between humans and social robots that respects diversity in experience, culture and language \cite{Nias}. In their paper they focus on education and schooling, but we think this statement could be generalized. Obremski et al. discuss the representation of multiple cultural backgrounds in one socially interactive agent and how an agent is perceived speaking in a none-native accent \cite{Obremski}. They partly observed a transfer of stereotypes and a difference in the perception of the robot.

Personality can be manifested in voice and language and according to the principle of similarity-attraction a person tends to like others with similar patterns. This hypothesis is explored by Pazylbekov et al. who let humans choose between robots with different dialectal features of the language \cite{Pazylbekov}. They suggest that dialect and language patterns are important in improving people's sympathy for robots.

Lugrin et al. investigated the effects of a regular language compared to High German \cite{Lugrin}. They used recorded human voices with a robot and had the participants listen to the robot. The text was recorded in three different speech styles: High German, Franconian accent and Franconian dialect. They expected effects on perceptions of competence, social skills and likability. However, these effects depended on whether the participants themselves spoke dialect or not.

\section{HUMANOID ROBOT PEPPER}

\begin{figure}
  \includegraphics[width=\textwidth]{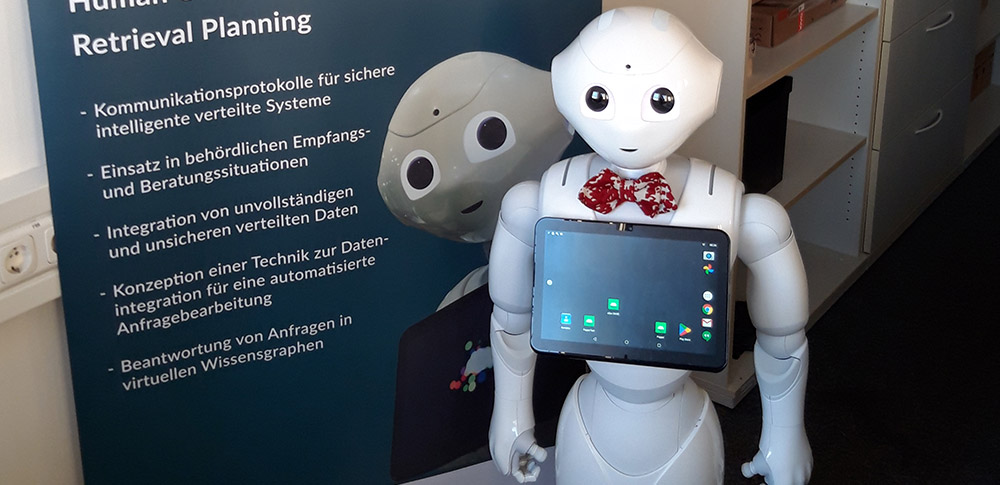}
  \caption{Humanoid Robot Pepper}
  \label{fig_pepper}
\end{figure}

The social humanoid robot Pepper, shown in Figure~\ref{fig_pepper}, was developed by Aldebaran and first released in 2015 \cite{Pepper}. The robot is 120 centimeters tall and optimized for human interaction. It is able to engage with people through conversation, gestures and its touch screen. Pepper is equipped with internal sensors, four directional microphones in his head and speakers for voice output. Speech recognition and dialogue is available in 15 languages.

Since research has generally shown that anthropomorphism improves social interaction \cite{Fink}, a humanoid social robot like Pepper is a good choice for HRI experiments. A human-like face and body language, the use of voice and a name of one's own are considered beneficial to the human relationship with the robot.

The robot features an open and fully programmable platform so that developers can program their own applications to run on Pepper using software development kits (SDKs) for programming languages like C++, Python or Java respectively Kotlin. This approach allows the development of robot applications for a wide variety of scenarios in a development environment familiar to most developers.

\begin{figure}
  \includegraphics[width=\textwidth]{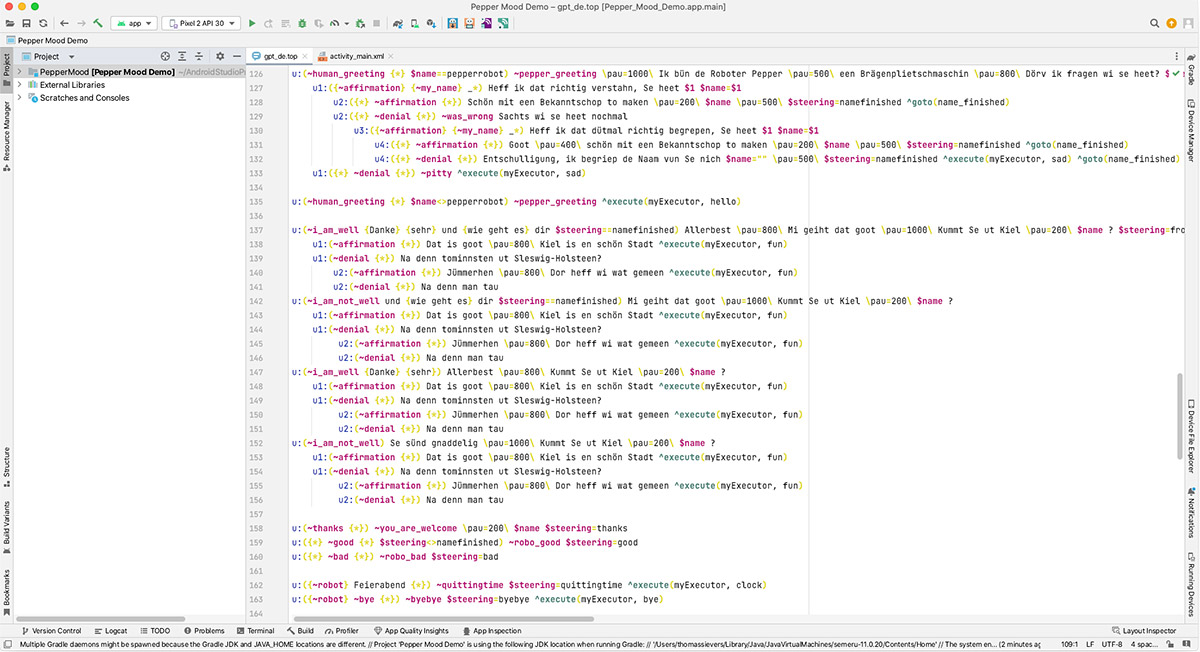}
  \caption{Dialog topic file}
  \label{fig_topic}
\end{figure}

Pepper's German language package and built-in speech recognition is used for understanding and talking with the user. When speaking in Low German, Pepper's synthesized voice, though not perfect, is clearly recognizable as speaking Low German. For robotic conversation we created topics, which are a set of rules, concepts and proposals. These elements, written in QiChat syntax as shown in Figure~\ref{fig_topic}, set the template for the conversation flow.

\section{THE SURVEY}

To measure possible effects that the two different ways of robot talk (High German vs. Low German) have on a human interlocutor's judgement of social attributes of the robot we use the 18-item Robotic Social Attributes Scale (RoSAS) \cite{Carpinella}. It comprises three underlying scale dimensions -- the factors \textit{warmth}, \textit{competence} and \textit{discomfort}. RoSAS is based on items from the Godspeed Scale \cite{Bartneck} and psychological literature on social perception. It is considered to be a psychologically valid scale of robotic perception.

We assume that a robot speaking Low German in an area with a corresponding cultural identity should make a difference in how it is perceived by human interlocutors. If so, the inclusion of cultural identity traits could make a difference in personality perceptions of a social robot.

\subsection{Experimental Setup}

We used a within-subject design where all participants had a conversation with the robot both in Low German and High German. The 17 participants in our study came from different walks of life and were aged between 19 and 61. We had 9 female and 8 male participants. To eliminate ordering effects we mixed the order of Low and High German for different participants. Thus, nearly half of the participants first spoke to the robot in Low German, followed by an interaction in which the robot used High German. For the other group, it was the other way around. Apart from the order, the experimental scenario was identical for all participants. The topic, content and course of the dialogue were also the same for everyone. The robot started the conversation with a greeting, asked for the name of its counterpart and then asked questions on various topics, mostly closed questions to which the human interlocutor could express agreement or disagreement. The name given by the human was repeatedly incorporated into the robot's speech in order to create a more natural conversation. Participants could also ask a few predefined questions.

Before interacting with the robot, each participant filled out a questionnaire that included questions about gender, age, place of origin and knowledge of Low German. All participants in the survey lived in the northern part of Germany and were more or less familiar with Low German, but most of them did not actively speak the language. We had 7 participants who rated their knowledge of Low German as good to very good. The other 10 participants had little or no knowledge of this regional language, but could mostly guess the meaning. In any case, the participants had to speak High German because our Pepper robot did not understand Low German.

After each dialogue session with the robot, the 18 items of the RoSAS were presented to the participants. They were asked, \textit{“Using the scale provided, how closely are the words below associated with your perception of the robot?”}. The participants responded using a 5-point likert scale from 1 = \textit{does not apply at all} to 5 = \textit{applies}. Every RoSAS factor comprises six items. For the factor \textit{warmth} they are: happy, feeling, social, organic, compassionate and emotional. \textit{Competence} includes: capable, responsive, interactive, reliable, competent and knowledgable. And \textit{discomfort} comprises: scary, strange, awkward, dangerous, awful and aggressive. Each factor consisting of six items could therefore receive a total of between 6 and 30 points, with a score of 6 representing the lowest agreement and 30 the highest.

\subsection{Results}

\begin{figure}
  \includegraphics[width=\textwidth]{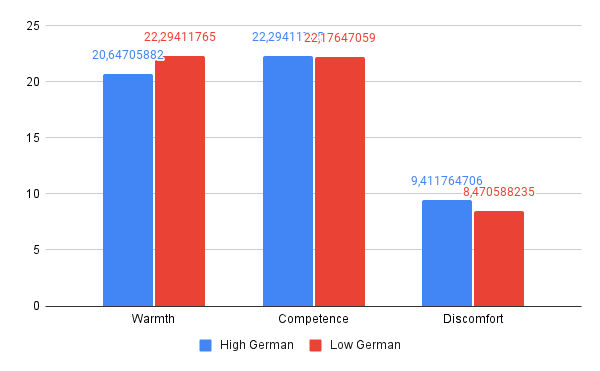}
  \caption{The results of the perceived warmth, competence and discomfort of the High German or Low German speaking Pepper Robot.}
  \label{fig_chart}
\end{figure}

We used a t-Test to determine whether there is a statistically significant difference between the perception of the conversation in High German versus the Low German variant in the mean. Figure~\ref{fig_chart} illustrates our results. In general, it can be seen that perceived warmth and competence tend to be high and perceived discomfort tends to be low.

\begin{figure}
  \includegraphics[width=\textwidth]{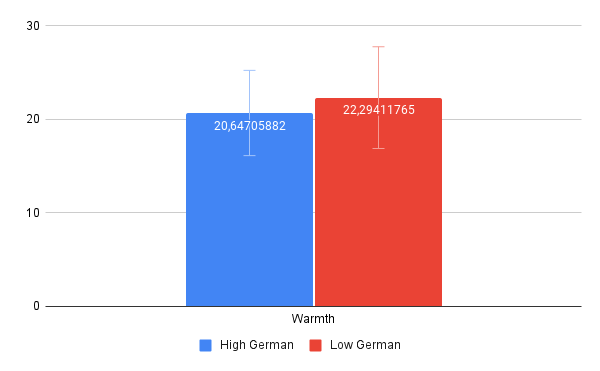}
  \caption{The results of the perceived warmth of the High German or Low German speaking Pepper Robot. Error bars show standard deviations.}
  \label{fig_warmth}
\end{figure}

\begin{figure}
  \includegraphics[width=\textwidth]{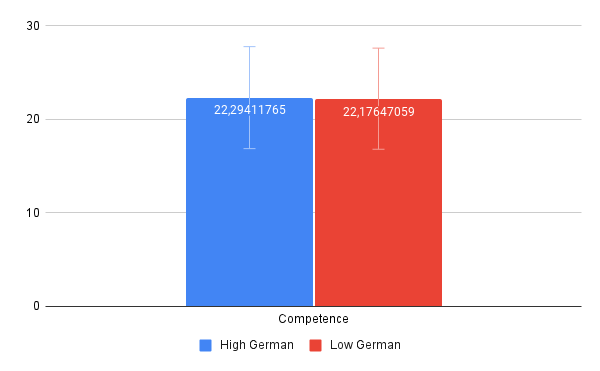}
  \caption{The results of the perceived competence of the High German or Low German speaking Pepper Robot. Error bars show standard deviations.}
  \label{fig_competence}
\end{figure}

\begin{figure}
  \includegraphics[width=\textwidth]{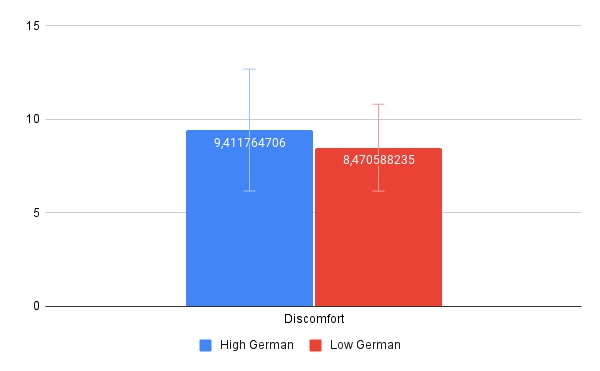}
  \caption{The results of the perceived discomfort of the High German or Low German speaking Pepper Robot. Error bars show standard deviations.}
  \label{fig_discomfort}
\end{figure}

The test showed in detail that participants perceived significantly higher warmth in the Low German version of the conversation (p = 0.027). Figure~\ref{fig_warmth} illustrates the result of the perceived warmth. For the competence factor, there was hardly any difference in perception (p = 0.378, Figure~\ref{fig_competence}). Finally, discomfort was perceived slightly less in the Low German version, but this cannot be considered a statistically significant result, as the p-value of p = 0.056 is somewhat above the significance level of 0.05. Figure~\ref{fig_discomfort} illustrates this result. But in our opinion, this result can still be understood as an indication of an influence on the factor discomfort. Table \ref{tbl:rosastable} shows the results as mean values for the 18 RoSAS items divided into areas of the three factors \textit{warmth}, \textit{competence} and \textit{discomfort}.

\begin{table}
  \centering
  \caption{RoSAS items and their mean values for High German and Low German}
  %\begin{tabular}{lll}
  %\begin{tabular}{p{0.25\textwidth}|p{0.25\textwidth}|p{0.25\textwidth}}
  \begin{tabular}[h]{c|l|c|c}
    & \textbf{RoSAS item}  & \textbf{High German} & \textbf{Low German} \\
    \hline
    & Happy & 4.235 & 4.412 \\
    & Feeling & 3.471 & 3.529 \\
    \raisebox{-.5\normalbaselineskip}[0pt][0pt]{\rotatebox[origin=c]{90}{Warmth}} & Social & 3.588 & 3.941 \\
    & Organic & 3.000 & 3.294 \\
    &Compassionate & 3.059 & 3.471 \\
    & Emotional & 3.294 & 3.647 \\
    \hline
    & Capable & 3.588 & 3.706 \\
    & Responsive & 4.000 & 3.824 \\
    \raisebox{-.5\normalbaselineskip}[0pt][0pt]{\rotatebox[origin=c]{90}{Competence}} & Interactive & 3.941 & 4.176 \\
    & Reliable & 3.588 & 3.412 \\
    & Competent & 3.882 & 3.765 \\
    & Knowledgable & 3.294 & 3.294 \\
    \hline
    & Scary & 1.588 & 1.353 \\
    & Strange & 2.353 & 2.176 \\
    \raisebox{-.5\normalbaselineskip}[0pt][0pt]{\rotatebox[origin=c]{90}{Discomfort}} & Awkward & 2.176 & 2.118 \\
    & Dangerous & 1.176 & 1.000 \\
    & Awful & 1.118 & 1.000 \\
    & Aggressive & 1.000 & 1.000 \\
  \end{tabular}
  \label{tbl:rosastable}
% Verweis im Text mittels \ref{tbl:rosastable}
\end{table}

\subsection{Discussion}

For our study, we used a real robot that engages in dialogue with humans using a synthetic voice, while many similar studies use virtual robots, pictures of robots or a robot saying something on video with pre-recorded human voice. This certainly has an impact on the perception of the conversation because the physical presence of the robot has a much more immediate effect.

Some of the participants were not used to the Low German language and, although they could mostly guess the meaning of the words, did not speak it at all. Some knew Low German as the language of their grandparents from their childhood. Perhaps that is one of the reasons why it is associated more with warmth or perhaps coziness than High German. As suggested in previous work, a regional language seems to increase perceived social skills as well as likeability and could reduce people's discomfort. It is perhaps somewhat surprising that the use of a regional language in our experiment had no effect on the robot's perception of competence, as Lugrin et al. mentioned previous work indicating a possible decrease in perceived competence with non-standard language \cite{Lugrin}.

\section{CONCLUSIONS AND FUTURE WORK}

In this paper, we present the results of a survey that asked whether a social robot speaking in a regional language is perceived differently in conversation in terms of warmth, competence and discomfort than when using the standard language. Our results suggest that the inclusion of cultural regional aspects, as found in languages, has a positive impact on the perceived personality of the robot, especially in terms of perceived warmth and discomfort in conversations.

In future work, the robot should also understand Low German, so that a dialogue is also possible with people who speak in this regional language. Gestures and body language as an essential part of expression should, if possible, also be adapted to regional characteristics and examined in their effect within the framework of a cultural identity.

\bibliographystyle{splncs04}

%\onecolumn
%\begin{figure}[h!]
%  \centering
%  \includegraphics[width=1.00\textwidth]{images/robot-emulator-chat.png}
%  \caption{Robot emulation viewer showing dialogue between robot and human with knowledge base answer, translation to easy language, and translation to Danish}
%  \label{fig_emulatorChat}
%\end{figure}
%\twocolumn

\end{document}